\documentclass{article}
\usepackage{iclr2027_conference,times}
\usepackage{amsmath}
\usepackage{amssymb}
\usepackage{xcolor}
\usepackage{graphicx}
\usepackage{booktabs}
\usepackage{multirow}
\usepackage{array}
\usepackage{hyperref}
\hypersetup{hidelinks}
\usepackage{url}
\usepackage{microtype}
\usepackage{pifont}

\AtBeginDocument{%
  \setlength{\abovedisplayskip}{5pt minus 2pt}%
  \setlength{\belowdisplayskip}{5pt minus 2pt}%
  \setlength{\abovedisplayshortskip}{2pt minus 1pt}%
  \setlength{\belowdisplayshortskip}{4pt minus 1pt}%
}

\newcommand{\cfg}[1]{\texttt{#1}}
\newcommand{\missing}{---}

\title{Looking One Step Ahead: An Anticipation-guided Agent for Object State Readout in Videos}

\author{
  \makebox[\dimexpr\textwidth-2\tabcolsep\relax][c]{
    \begin{tabular}{@{}c@{}}
      \textbf{Yuanhe Tian\textsuperscript{\ding{170}} \qquad
      Yan Song\textsuperscript{\ding{171}}\thanks{Corresponding author.}} \\[2pt]
      {\normalfont\small\textsuperscript{\ding{170}}Zhongguancun Academy \qquad
      \textsuperscript{\ding{171}}University of Science and Technology of China} \\
      {\normalfont\small\textsuperscript{\ding{170}}\texttt{yhtian94@gmail.com} \qquad
      \textsuperscript{\ding{171}}\texttt{clksong@gmail.com}}
    \end{tabular}
  }
}
\iclrfinalcopy

\begin{document}

\maketitle
\fancyhead{}
\begin{abstract}
Understanding the progress of an activity requires tracking how the objects involved change over time. 
Object state readout provides this information by describing a target object's states and transitions across video frames. 
Although multimodal large language models (MLLMs) generate detailed visual descriptions, these descriptions may leave the attributes needed to distinguish successive states insufficiently specified. 
In this paper, we propose look-ahead state readout (LASER), an anticipation-guided agent for object state readout that uses possible next-step changes to determine which attributes to inspect in the present. 
Given temporally ordered frames and a target object, LASER converts anticipated changes into targeted visual questions and inspects the current frame to establish attribute values. 
It then checks state readouts and forecasts against independent observations of subsequent frames, refining a state record used to generate the final description. 
Reusable textual observation skills guide this process and are refined by tracing discrepancies with reference captions to the responsible instructions. 
Candidate revisions are selected on separate videos based on improved caption quality without increased contradictions with reference facts. 
Model parameters remain frozen throughout adaptation and inference, and the revised skills remain fixed during evaluation. 
Experiment results and further analysis on OSCaR and PLM RCap show consistent improvements of LASER over strong baselines and existing approaches, which demonstrates the effectiveness of our approach.\footnote{Code is available at \url{https://github.com/synlp/LASER}.}

\end{abstract}

\section{Introduction}
\label{sec:introduction}
Object state readout describes how a specified object's condition changes across visual observations, providing a language account of its evolving state \citep{nguyen2024oscar,cho2025perceptionlm}.
Whereas image captioning describes the content of a single image, state readout tracks object attributes across time to distinguish successive states and their transitions.
This makes state readout an important task for procedural planning and robotic interaction, where understanding how actions change objects supports subsequent decisions \citep{nguyen2024oscar}.
The central challenge is deciding which attributes reveal a meaningful change, rather than merely recognizing the object or naming its activity.

Multimodal large language models (MLLMs) provide a foundation for describing visual observations in natural language.
Recent studies improve their video descriptions through detailed spatial and temporal grounding \citep{xue2025progressaware,cho2025perceptionlm,kazakos2025largescale,athar2025vicas}, while temporal benchmarks evaluate their understanding of motion and event order \citep{liu2024tempcompass,cai2024temporalbench}.
Related agent studies explore anticipation-guided observation \citep{yang2026streamagent} and feedback-driven refinement of reusable instructions \citep{zhao2024expel,zhang2026agentic}.
These capabilities support object state readout, but accurate grounding and temporal ordering alone do not specify which attributes a state description needs to retain for subsequent comparison.
A description may correctly identify an object and its surrounding activity while leaving the property that changes insufficiently specified, weakening the basis for describing its transition across images.
Selecting these attributes requires considering which properties distinguish the object's possible subsequent states.
For example, for a drawer, opening, closing, and remaining unchanged make its current extension relevant, even when that detail receives little emphasis in a general scene description.
Anticipated changes thus provide a basis for deciding what to inspect, while visual evidence establishes the current attribute value.
The next image provides further evidence for assessing those changes and identifying attributes that require closer inspection.
This connection between possible changes and observable properties motivates using anticipation for attribute selection, subsequent evidence to refine state descriptions, and feedback on those descriptions to improve reusable observation instructions.

\begin{figure}[t]
\centering
\includegraphics[width=\linewidth, trim=0 40 0 0]{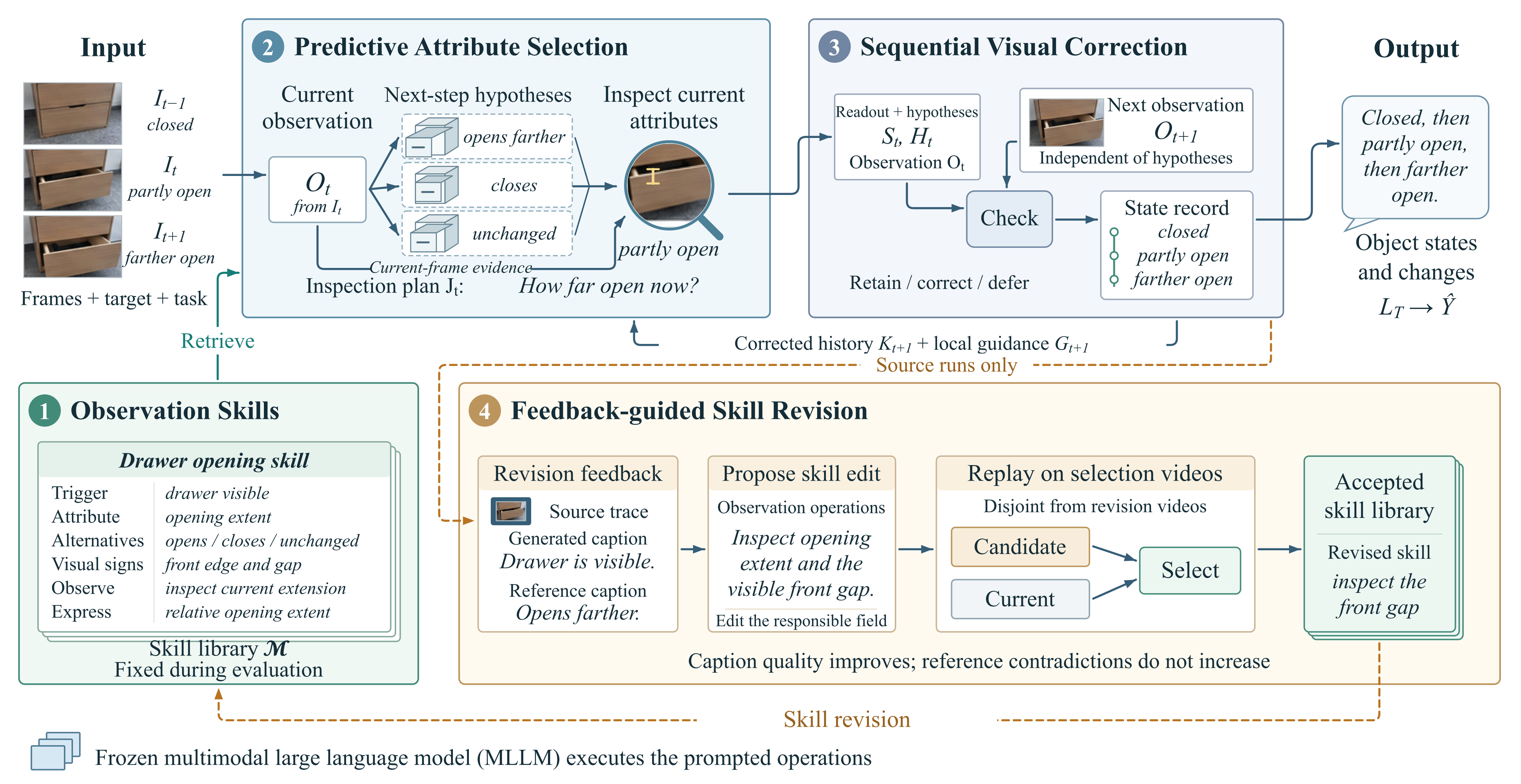}
\caption{Overview of LASER. (1) Observation skills guide (2) predictive attribute selection to anticipate possible changes and inspect current attributes. (3) Sequential visual correction checks the readout and hypotheses against the next observation, obtained independently of the hypotheses. Corrected history and local guidance support subsequent inspection; the accumulated state record supports the final description. (4) Feedback-guided skill revision uses source-video traces and reference captions to propose edits, then selects edits on separate selection videos when caption quality improves without increasing reference contradictions. Solid arrows indicate information flow; orange dashed arrows indicate source-trace transfer and skill-library updates for adaptation. The MLLM remains frozen, and the skill library is fixed during evaluation.}
\label{fig:approach}
\vspace{-0.3cm}
\end{figure}

In this paper, we propose look-ahead state readout (LASER), an anticipation-guided agent that guides an MLLM to inspect an object's current state by anticipating possible next-step changes.
Given a sequence of video frames and a target object, LASER uses observation skills, expressed as reusable textual instructions, to anticipate alternative changes and identify the attributes that distinguish them.
Predictive attribute selection turns these attributes into visual questions and examines the current image to establish their values.
Sequential visual correction then checks the resulting descriptions and anticipated changes against subsequent images, maintaining an evidence-supported state record for the final description.
To improve the observation skills, feedback-guided skill revision traces discrepancies between generated and reference captions to the instructions responsible for observation and description.
It proposes focused edits and retains those that improve caption quality on separate source videos without increasing contradictions with the references.
This process adapts observation instructions while keeping all model parameters fixed, and the resulting skills remain fixed during evaluation.
We evaluate LASER on English benchmarks across various MLLMs, where our approach outperforms strong baselines and existing studies, demonstrating the effectiveness of our approach.

\section{The Approach}
\label{sec:approach}

We propose look-ahead state readout (LASER), an anticipation-guided agent that considers possible next-step changes to guide observation of an object's current state.
Figure \ref{fig:approach} presents its overall architecture, comprising four components: \textit{observation skills}, \textit{predictive attribute selection}, \textit{sequential visual correction}, and \textit{feedback-guided skill revision}.
LASER organizes observation and reasoning operations and maintains the state record and local guidance; a frozen MLLM executes the visual inspection and language reasoning within these operations.
Specifically, given $T$ temporally ordered video frames $V=(I_1,\ldots,I_T)$, a target prompt $P$ identifying the object, and a task instruction $\mathcal{T}$ specifying the description scope, LASER processes the observations in sequence.
For each observation $I_t$, with $t=1,\ldots,T$, LASER retrieves \textit{observation skills} from the skill library $\mathcal{M}$ to guide which attributes of the target object to inspect and how to describe them.
\textit{Predictive attribute selection} uses these skills to propose possible next-step changes, converts their distinguishing attributes into visual questions, and checks the current image to establish the attribute values.
When the next observation arrives, \textit{sequential visual correction} checks earlier descriptions and anticipated changes against the new evidence. It updates a state record $L_t$ of observed attributes, changes, and supporting evidence, and local guidance $G_t$ containing unresolved questions for subsequent inspection.
After the final observation, LASER composes the description $\widehat{Y}$ from the final state record $L=L_T$.
During adaptation, \textit{feedback-guided skill revision} uses discrepancies in these readout sequences and reference captions to propose instruction edits, then selects edits through replay on separate videos.

The observation cycle in Figure \ref{fig:approach} follows readout at $I_t$ through correction when $I_{t+1}$ arrives. The feedback path returns corrected history and local guidance to skill selection and current-attribute inspection for that next observation.
LASER returns the description $\widehat{Y}$ and its supporting state record $L$:
\begin{equation}
(\widehat{Y},L)=\operatorname{LASER}(V,P,\mathcal{T};\mathcal{M})
\label{eq:laser}
\end{equation}
The following subsections describe observation skills, predictive attribute selection, sequential visual correction, and feedback-guided skill revision, respectively.

\subsection{Observation Skills}
\label{sec:skills}

An observation skill is a structured textual instruction that tells the agent what to inspect about an object and how to describe the visual evidence.
Each skill contains six fields: a trigger specifying when the instruction applies, an attribute scope identifying what to inspect, alternative changes suggesting possible next states, visual signs distinguishing those alternatives, observation operations specifying how to check the current image, and expression constraints governing the resulting description.
We organize these observation instructions into a skill library $\mathcal{M}$.
For each observation, the MLLM selects relevant skills to guide inspection (see Section \ref{sec:prediction} for details).
Feedback-guided skill revision updates these instructions using caption feedback (Section \ref{sec:revision}).

\subsection{Predictive Attribute Selection}
\label{sec:prediction}

Predictive attribute selection uses possible next-step changes to determine which attributes matter for describing the object's current state.
For each image $I_t$, the MLLM first observes the target using the task instruction $\mathcal{T}$ and the target information available up to that observation, denoted by $P_{\leq t}$.
It produces an observation description $O_t$ containing visible attributes and their values, supporting image regions, and any uncertainty.
This initial observation receives no predicted changes, so its description is grounded in the image.
For $t>1$, LASER checks earlier descriptions against the new observation before selecting skills.
The resulting $K_t$ contains the earlier states and observed changes after evidence-supported corrections, while local guidance $G_t$ records questions that still require inspection.
Both are empty for the first observation.
The MLLM selects skills whose triggers and attribute scopes match $O_t$, $\mathcal{T}$, $K_t$, and $G_t$, forming $\mathcal{M}_t$.
When no skill matches, the MLLM follows the task instruction and directly inspects observable attributes of the target object.

Using the selected skills, the MLLM anticipates how the target might change by the next observation.
For $t<T$, it generates a set of hypotheses $H_t$ from the current description and the information accumulated so far:
\begin{equation}
H_t=\operatorname{Anticipate}(O_t,K_t,G_t,\mathcal{M}_t,\mathcal{T})
\label{eq:hypotheses}
\end{equation}
Each hypothesis identifies an object attribute, a possible next value or change direction, and a visual sign that would distinguish that outcome.
For a drawer, alternatives might be opening further, closing, or remaining unchanged; they make its current extension an informative attribute to inspect.
These hypotheses concern the next available observation and use only the images observed so far.

The MLLM converts the hypotheses into an inspection plan $J_t$: a set of questions about the target's current attributes.
Each question specifies what visual evidence to examine and how that evidence supports a description.
Questions remain neutral about the expected answer; for example, the drawer hypothesis leads to asking how far the drawer is currently extended.
The plan also includes task-relevant attributes already identified in $O_t$ and unresolved questions in $G_t$.
We denote the hypotheses available to planning by $H_t^{\mathrm{use}}$: these are $H_t$ for $t<T$ and the preceding hypotheses $H_{T-1}$ for the final observation.
The plan is constructed as:
\begin{equation}
J_t=\operatorname{Plan}(O_t,H_t^{\mathrm{use}},\mathcal{M}_t,G_t,\mathcal{T})
\label{eq:attributes}
\end{equation}
The MLLM inspects the current image according to $J_t$ and produces the state readout $S_t$ by $S_t=\operatorname{Read}(I_t,P_{\leq t},J_t,O_t)$.
The readout records attribute values and their supporting visual evidence, leaving values unresolved when the image is insufficient.
Subsequent observations check this readout and its associated hypotheses through sequential visual correction.

\subsection{Sequential Visual Correction}
\label{sec:correction}
When the agent processes image $I_t$ ($t>1$), sequential visual correction checks the preceding state description $S_{t-1}$ and predicted changes $H_{t-1}$ against the new visual evidence.
The MLLM uses the observation descriptions $O_{t-1}$ and $O_t$, which are obtained without predicted values.
If these descriptions provide insufficient evidence, the MLLM examines the two images again to answer the unresolved visual questions.
This additional inspection produces $W_t$ without receiving the preceding claimed value or predicted outcome; otherwise, $W_t$ is empty.
The comparison produces feedback $E_t$:
\begin{equation}
E_t=\operatorname{Compare}(S_{t-1},H_{t-1},O_{t-1},O_t,W_t,\mathcal{T})
\label{eq:feedback}
\end{equation}
The feedback identifies supported or contradicted hypotheses, observed state changes, errors in earlier descriptions, and questions that the available evidence does not resolve.
If an object previously observed as standing is now seated, the agent records the transition and rejects a prediction of continued standing; it does not change the earlier standing observation.
An earlier description is corrected only when visual evidence contradicts that description.
When the object's identity across the two images is uncertain, the agent leaves the transition unresolved.

The state record and local guidance start empty: $L_0=\varnothing$ and $G_0=\varnothing$.
For $t>1$, the correction operation applies evidence-supported repairs to the preceding state record $L_{t-1}$, producing $K_t$, while the guidance operation updates the unresolved questions in $G_t$:
\begin{equation}
K_t=\operatorname{Correct}(L_{t-1},E_t),
\quad
G_t=\operatorname{Guide}(G_{t-1},E_t)
\label{eq:correctedprefix}
\end{equation}
Here, $K_t$ contains the corrected history before the current state is added.
Predictive attribute selection then uses $K_t$ and $G_t$ to guide inspection of the current image, yielding $S_t$.
The agent adds this readout to the corrected history to obtain the updated state record:
\begin{equation}
L_t=\operatorname{Update}(K_t,S_t)
\label{eq:record}
\end{equation}
For the first image, $E_1$, $K_1$, and $G_1$ are empty, and $S_1$ initializes the state record.

After all $T$ images have been processed, the MLLM generates the final description from $L_T$.
It selects supported states and changes within the object and temporal scope specified by $\mathcal{T}$:
\begin{equation}
\widehat{Y}=\operatorname{Compose}(L_T,\mathcal{T})
\label{eq:compose}
\end{equation}

\subsection{Feedback-guided Skill Revision}
\label{sec:revision}
Feedback-guided skill revision improves observation instructions using existing reference captions while keeping model parameters fixed.
We divide the source videos into a revision set $\mathcal{D}_{\mathrm{rev}}$ for identifying problems and proposing edits, and a selection set $\mathcal{D}_{\mathrm{sel}}$ for evaluating those edits.
The sets contain different videos to test whether edits generalize beyond the examples that motivate them.
In each round, LASER processes the revision videos with the current library.
The MLLM compares generated captions with references and examines the corresponding observations, hypotheses, inspection questions, and state records to identify discrepancy sources.
The comparison identifies omissions and explicit contradictions; details unmentioned in the reference remain unjudged.

When a relevant attribute is never inspected, it examines whether the skill's trigger or alternative changes fail to direct attention to that attribute.
When inspection produces an unsupported value, it examines the visual signs and observation operations.
When a supported fact appears in the state record but is missing from the final caption, it examines the expression constraints.
For a discrepancy linked to an existing skill, the MLLM proposes an edit to one responsible field; if no skill covers the situation, it proposes a new skill.
Each proposal produces a candidate library $\mathcal{M}'$ derived independently from the current library $\mathcal{M}$.
The MLLM proposes no edit when the available evidence does not identify a responsible instruction.

LASER compares each candidate with the current library on $\mathcal{D}_{\mathrm{sel}}$.
For an example $x$, which contains the images, target information, and task instruction, let $Y_x^*$ denote its reference caption.
The score $q$ measures agreement between the generated and reference captions, giving the mean caption quality:
\begin{equation}
Q(\mathcal{M})=\frac{1}{|\mathcal{D}_{\mathrm{sel}}|}\sum_{x\in\mathcal{D}_{\mathrm{sel}}}q\bigl(\operatorname{Caption}(x;\mathcal{M}),Y_x^*\bigr)
\label{eq:quality}
\end{equation}
Here, $\operatorname{Caption}$ returns the description produced by Equation \ref{eq:laser}; references are used for scoring rather than caption generation.
To check contradictions, the MLLM extracts explicit facts from each reference caption once and uses the same facts to assess every candidate.
An MLLM judge determines whether a generated caption contains a claim incompatible with a reference fact concerning the same object and temporal scope.
Omissions, details absent from the reference, and uncertain comparisons do not count as contradictions.
The contradiction score $X(\mathcal{M})$ is the fraction of selection examples containing at least one explicit contradiction.

Selection requires an improvement in caption quality without increasing reference contradictions.
For a positive quality margin $\epsilon$, the acceptance rule is:
\begin{equation}
\operatorname{Accept}(\mathcal{M}')=\mathbf{1}\!\left[Q(\mathcal{M}')-Q(\mathcal{M})\geq\epsilon,\;X(\mathcal{M}')\leq X(\mathcal{M})\right]
\label{eq:accept}
\end{equation}
The accepted candidate with the highest caption quality replaces the current library; if none is accepted, the library is retained.
The resulting skill library remains fixed during evaluation.

\section{Experiment Setting}
\label{sec:experiments}

\subsection{Datasets}
\label{sec:datasets}
We evaluate object-conditioned captioning on OSCaR \citep{nguyen2024oscar} and PLM RCap \citep{cho2025perceptionlm}.
We use OSCaR's released EPIC state-change slot, containing 7K training clips and 500 test clips.
Each example pairs three ordered frames with a named object and a reference description of its state change.
PLM RCap contains 179K training, 4K development, and 10K test region--interval captions.
Given a target masklet and its video interval, the task describes the target's actions and interactions.
For each benchmark, we select 500 examples from its training split and use their reference captions to adapt observation skills.
We divide these examples into video-disjoint revision and selection sets, using the former to propose instruction edits and the latter to evaluate them.

\subsection{Baselines and Comparing Models}
\label{sec:baselines}

We compare LASER with the following baselines.
\textbf{Direct Captioning} generates a target description from the input frames and task instruction in one pass.
\textbf{Sequential Readout} describes object states in temporal order and summarizes their changes without anticipating future states.
\textbf{Generic Reinspection} extends sequential readout by checking for omissions or inconsistencies.
\textbf{Attribute Checklist} selects current-state attributes for inspection and retains visual correction, without next-step hypotheses.
\textbf{Free Anticipation} predicts possible future events without identifying their distinguishing attributes and visual signs.
\textbf{LASER (No Skill Revision)} uses unadapted initial observation skills.
\textbf{LASER (GEPA Optimization)} uses GEPA \citep{agrawal2026gepa} to optimize the same prompt interfaces instead of feedback-guided skill revision.
\textbf{LASER (Full)} is our full approach.

We also compare LASER with five existing approaches.
\textbf{InstanceCap} \citep{fan2025instancecap} organizes video instances into structured descriptions of their appearance, actions, motion, and position.
\textbf{Matters} \citep{peng2025patch} describes local image patches and hierarchically aggregates their details through semantic filtering.
\textbf{RICO} \citep{wang2025rico} reconstructs images from captions and revises the captions by comparing the reconstructions with the original images.
\textbf{ReAgent-V} \citep{zhou2026reagent} combines task-relevant frame selection and tool-derived evidence with critical feedback and multi-perspective answer revision.
\textbf{ScaleCap} \citep{xing2026scalecap} expands captions through targeted visual questions and filters unsupported sentences through contrastive scoring.
We adapt these approaches to object-conditioned captioning and evaluate them on the benchmarks.

\begin{table}[t]
\centering
\renewcommand{\missing}{---}
\renewcommand{\cfg}[1]{\texttt{#1}}
\small
\setlength{\tabcolsep}{3pt}
\renewcommand{\arraystretch}{1.12}
\setlength{\abovecaptionskip}{0pt}
\setlength{\belowcaptionskip}{6pt}
\caption{Comparison of our approach with various baselines across four backbone MLLMs on the test set of OSCaR and PLM RCap, where ``+ LASER (full)'' denotes our full model.}
\label{tab:main}
\begin{tabular}{@{}>{\raggedright\arraybackslash}p{0.36\linewidth}|*{4}{>{\centering\arraybackslash}p{\dimexpr(\linewidth-0.36\linewidth-10\tabcolsep-2\arrayrulewidth)/5\relax}}|>{\centering\arraybackslash}p{\dimexpr(\linewidth-0.36\linewidth-10\tabcolsep-2\arrayrulewidth)/5\relax}@{}}
\toprule
\multirow{2}{*}{\textbf{Approach}} & \multicolumn{4}{c|}{\textbf{OSCaR}} & \textbf{PLM RCap} \\
\cmidrule(lr){2-5}\cmidrule(lr){6-6}
 & BLEU$\uparrow$ & ROUGE-1$\uparrow$ & ROUGE-2$\uparrow$ & ROUGE-L$\uparrow$ & Score$\uparrow$ \\
\midrule
Qwen3.5-4B & 7.7 & 28.4 & 8.7 & 25.2 & 24.7 \\
\quad + Sequential Readout & 8.0 & 29.2 & 8.8 & 26.2 & 26.1 \\
\quad + Generic Reinspection & 8.0 & 29.8 & 8.7 & 25.9 & 25.3 \\
\quad + Attribute Checklist & 8.6 & 29.6 & 9.2 & 26.7 & 26.8 \\
\quad + Free Anticipation & 8.1 & 29.9 & 9.0 & 25.2 & 25.3 \\
\quad + LASER (No Skill Revision) & 8.7 & 30.1 & 8.9 & 26.7 & 25.9 \\
\quad + LASER (GEPA Optimization) & 9.0 & \textbf{31.1} & 9.1 & \textbf{29.0} & \textbf{27.5} \\
\quad \textbf{+ LASER (Full)} & \textbf{9.3} & 30.9 & \textbf{9.3} & 28.9 & 27.3 \\
\midrule
Qwen3.5-9B & 8.6 & 29.8 & 9.5 & 26.9 & 25.6 \\
\quad + Sequential Readout & 9.1 & 30.6 & 9.5 & 28.5 & 26.7 \\
\quad + Generic Reinspection & 9.3 & 31.1 & 9.9 & 28.7 & 25.8 \\
\quad + Attribute Checklist & 9.6 & 30.8 & 9.9 & 27.7 & 27.2 \\
\quad + Free Anticipation & 9.0 & 30.5 & 9.6 & 28.5 & 26.4 \\
\quad + LASER (No Skill Revision) & 9.1 & 30.5 & 9.8 & 28.2 & 26.8 \\
\quad + LASER (GEPA Optimization) & 9.5 & 31.6 & \textbf{10.1} & \textbf{29.9} & 27.5 \\
\quad \textbf{+ LASER (Full)} & \textbf{9.9} & \textbf{32.0} & \textbf{10.1} & 29.6 & \textbf{28.1} \\
\midrule
Qwen3.5-35B-A3B & 8.8 & 30.4 & 9.7 & 27.7 & 26.3 \\
\quad + Sequential Readout & 9.0 & 30.3 & 10.0 & 27.9 & 27.5 \\
\quad + Generic Reinspection & 9.0 & 31.0 & 9.8 & 29.4 & 27.0 \\
\quad + Attribute Checklist & 9.6 & 31.5 & 10.1 & 29.3 & 27.3 \\
\quad + Free Anticipation & 9.2 & 31.0 & 9.7 & 28.2 & 27.3 \\
\quad + LASER (No Skill Revision) & 9.7 & 30.9 & 10.0 & 28.9 & 28.1 \\
\quad + LASER (GEPA Optimization) & 10.3 & 32.2 & 10.7 & \textbf{30.7} & \textbf{29.0} \\
\quad \textbf{+ LASER (Full)} & \textbf{10.6} & \textbf{33.1} & \textbf{10.8} & 30.4 & 28.8 \\
\midrule
Ministral-3-3B & 6.8 & 27.1 & 6.1 & 24.0 & 22.9 \\
\quad + Sequential Readout & 7.6 & 28.2 & 6.0 & 24.9 & 23.8 \\
\quad + Generic Reinspection & 8.1 & 28.4 & 6.3 & 25.4 & 23.8 \\
\quad + Attribute Checklist & 8.1 & 28.6 & 7.0 & 25.5 & 23.6 \\
\quad + Free Anticipation & 7.4 & 27.7 & 7.4 & 25.4 & 23.4 \\
\quad + LASER (No Skill Revision) & 8.0 & 28.7 & 6.9 & 25.9 & 23.3 \\
\quad + LASER (GEPA Optimization) & 8.1 & 29.1 & 8.0 & 26.4 & 24.7 \\
\quad \textbf{+ LASER (Full)} & \textbf{8.8} & \textbf{29.9} & \textbf{8.2} & \textbf{27.0} & \textbf{25.2} \\
\bottomrule
\end{tabular}
\vspace{-0.3cm}
\end{table}

\subsection{Implementation Details}
\label{sec:implementation}

Our backbones MLLMs comprise Qwen3.5-4B, Qwen3.5-9B, and Qwen3.5-35B-A3B \citep{qwenteam2026qwen35}, plus Ministral-3-3B \citep{liu2026ministral}.
The two dense Qwen models each have 32 language layers, with hidden dimensions of 2,560 and 4,096 for the 4B and 9B variants, respectively.
The 35B-A3B variant uses 40 language layers with a hidden dimension of 2,048; its mixture-of-experts layers select 8 of 256 routed experts per token and include a shared expert, with an intermediate dimension of 512 for each expert.
Ministral-3-3B has 26 language layers with a hidden dimension of 3,072 and a 24-layer vision encoder.

We divide the adaptation examples into revision and selection sets at a 4:1 ratio, keeping examples from the same video in one set.
For PLM RCap, we uniformly sample eight frames from each target interval; OSCaR uses its three provided frames.
At each observation, LASER selects up to three relevant skills and generates up to three alternative next-step changes.
Skill adaptation runs for at most five rounds, with up to three candidate libraries evaluated per round.
We normalize caption quality scores to $[0,1]$ and require an improvement of at least $\epsilon=0.005$ without increasing reference contradictions.
Adaptation stops when no candidate satisfies these criteria.
All model parameters remain frozen, and the adapted skill library remains fixed during evaluation.

For evaluation, we use different metrics for the two benchmarks.
For OSCaR, we report BLEU \citep{papineni2002bleu}, ROUGE-1, ROUGE-2, and ROUGE-L \citep{lin2004rouge}, with ROUGE-L as the primary metric.
For PLM RCap, we use the official language-model-judged reference-caption similarity score \citep{cho2025perceptionlm}.
Higher values indicate better caption agreement on both tasks.
Evaluation references are accessible only to the scorer.

\section{Results and Analysis}
\label{sec:results}

\subsection{Overall Results}

Table \ref{tab:main} presents the results of our approach and baselines on OSCaR and PLM RCap across four backbone MLLMs.
There are several observations.
First, LASER consistently outperforms direct captioning, sequential readout, and generic reinspection across all metrics.
These gains suggest that accurate state descriptions benefit from selecting informative attributes beyond organizing observations temporally or inspecting them repeatedly.
Second, LASER achieves higher scores than both attribute checklists and free anticipation across the evaluated backbones.
This supports connecting possible changes to specific visual attributes, so that anticipation guides what to inspect in the current image.
Third, feedback-guided skill revision improves every reported metric over LASER with the initial skill library.
This indicates that caption feedback helps refine reusable observation instructions without updating model parameters.
Fourth, LASER improves over direct captioning for all four backbones, spanning different model sizes, model families, and dense and mixture-of-experts architectures, which demonstrates the robustness of our approach.

Table \ref{tab:comparison} compares LASER with five existing approaches adapted to object-conditioned captioning using Qwen3.5-9B as the backbone MLLM.
LASER outperforms all compared approaches on both benchmarks, which demonstrates the effectiveness of using anticipated changes to identify state-distinguishing attributes and refining their descriptions through subsequent visual evidence and caption feedback.

\begin{table}[t]
\centering
\renewcommand{\missing}{---}
\small
\setlength{\tabcolsep}{3pt}
\renewcommand{\arraystretch}{1.12}
\setlength{\abovecaptionskip}{0pt}
\setlength{\belowcaptionskip}{6pt}
\caption{Comparison of our approach with existing studies on using Qwen3.5-9B as the backbone MLLM. The best and second-best results are marked by boldface and underlines, respectively.}
\label{tab:comparison}
\begin{tabular}{@{}>{\raggedright\arraybackslash}p{0.36\linewidth}|*{4}{>{\centering\arraybackslash}p{\dimexpr(\linewidth-0.36\linewidth-10\tabcolsep-2\arrayrulewidth)/5\relax}}|>{\centering\arraybackslash}p{\dimexpr(\linewidth-0.36\linewidth-10\tabcolsep-2\arrayrulewidth)/5\relax}@{}}
\toprule
\multirow{2}{*}{\textbf{Approach}} & \multicolumn{4}{c|}{\textbf{OSCaR}} & \textbf{PLM RCap} \\
\cmidrule(lr){2-5}\cmidrule(lr){6-6}
 & BLEU$\uparrow$ & ROUGE-1$\uparrow$ & ROUGE-2$\uparrow$ & ROUGE-L$\uparrow$ & Score$\uparrow$ \\
\midrule
Qwen3.5-9B & 8.6 & 29.8 & 9.5 & 26.9 & 25.6 \\
InstanceCap \citep{fan2025instancecap} & \underline{9.6} & 31.1 & \underline{9.9} & 27.6 & 27.4 \\
Matters \citep{peng2025patch} & 9.1 & 30.2 & 9.4 & 27.1 & 25.4 \\
RICO \citep{wang2025rico} & 8.9 & 30.1 & \underline{9.9} & 27.5 & 25.5 \\
ReAgent-V \citep{zhou2026reagent} & 9.2 & \underline{31.4} & 9.8 & 28.7 & \underline{27.8} \\
ScaleCap \citep{xing2026scalecap} & 9.3 & 30.1 & 9.8 & \underline{29.4} & 26.6 \\
\textbf{LASER} & \textbf{9.9} & \textbf{32.0} & \textbf{10.1} & \textbf{29.6} & \textbf{28.1} \\
\bottomrule
\end{tabular}
\vspace{-0.3cm}
\end{table}

\subsection{Ablation Study}
We examine whether anticipated changes help identify current-state evidence and whether independent observation limits reliance on predicted values.
Using Qwen3.5-9B as the backbone MLLM on both benchmarks, we independently remove current reinspection, remove discriminating visual signs, or expose the initial observer to preceding hypotheses, retaining the full approach's revised skills in each case.
Table \ref{tab:attributes} compares these variants with full LASER and direct captioning.
Removing current reinspection lowers every metric, supporting the use of anticipated changes to guide a second examination of the present image.
Removing discriminating signs also lowers every metric and produces the largest ROUGE-L decline among the ablations.
This pattern suggests that identifying possible changes is more useful when those changes are connected to explicit visual criteria for inspection.
Exposing the initial observer to predicted changes reduces all reported scores and produces the largest decline in RCap similarity among the ablations.
The result supports keeping initial visual observations independent of forecasts, so that later comparisons rely on separately obtained evidence.
All three ablated variants remain above direct captioning, indicating that each removed operation contributes to the full approach while the remaining observation structure retains value.

\begin{table}[t]
\centering
\renewcommand{\missing}{---}
\renewcommand{\cfg}[1]{\texttt{#1}}
\small
\setlength{\tabcolsep}{3pt}
\renewcommand{\arraystretch}{1.15}
\setlength{\abovecaptionskip}{0pt}
\setlength{\belowcaptionskip}{6pt}
\caption{Results of ablation study of predictive observation with Qwen3.5-9B.}
\label{tab:attributes}
\begin{tabular}{@{}>{\raggedright\arraybackslash}p{0.36\linewidth}|>{\centering\arraybackslash}p{\dimexpr(\linewidth-0.36\linewidth-10\tabcolsep-2\arrayrulewidth)/5\relax}>{\centering\arraybackslash}p{\dimexpr(\linewidth-0.36\linewidth-10\tabcolsep-2\arrayrulewidth)/5\relax}>{\centering\arraybackslash}p{\dimexpr(\linewidth-0.36\linewidth-10\tabcolsep-2\arrayrulewidth)/5\relax}>{\centering\arraybackslash}p{\dimexpr(\linewidth-0.36\linewidth-10\tabcolsep-2\arrayrulewidth)/5\relax}|>{\centering\arraybackslash}p{\dimexpr(\linewidth-0.36\linewidth-10\tabcolsep-2\arrayrulewidth)/5\relax}@{}}
\toprule
\multirow{2}{*}{\textbf{Approach}} & \multicolumn{4}{c|}{\textbf{OSCaR}} & \textbf{PLM RCap} \\
\cmidrule(lr){2-5}\cmidrule(lr){6-6}
 & BLEU$\uparrow$ & ROUGE-1$\uparrow$ & ROUGE-2$\uparrow$ & ROUGE-L$\uparrow$ & Score$\uparrow$ \\
\midrule
\textbf{LASER (Full)} & \textbf{9.9} & \textbf{32.0} & \textbf{10.1} & \textbf{29.6} & \textbf{28.1} \\
\midrule
w/o Current Reinspection & 9.1 & 31.2 & 10.0 & 28.9 & 27.4 \\
w/o Discriminating Signs & 9.8 & 31.7 & 9.8 & 28.1 & 27.5 \\
w/o Independent Observation & 9.3 & 31.8 & 9.7 & 29.0 & 27.3 \\
\midrule
Qwen3.5-9B (Base) & 8.6 & 29.8 & 9.5 & 26.9 & 25.6 \\
\bottomrule
\end{tabular}
\end{table}

\begin{figure}[t]
\centering
\includegraphics[width=\linewidth, trim=0 100 0 0]{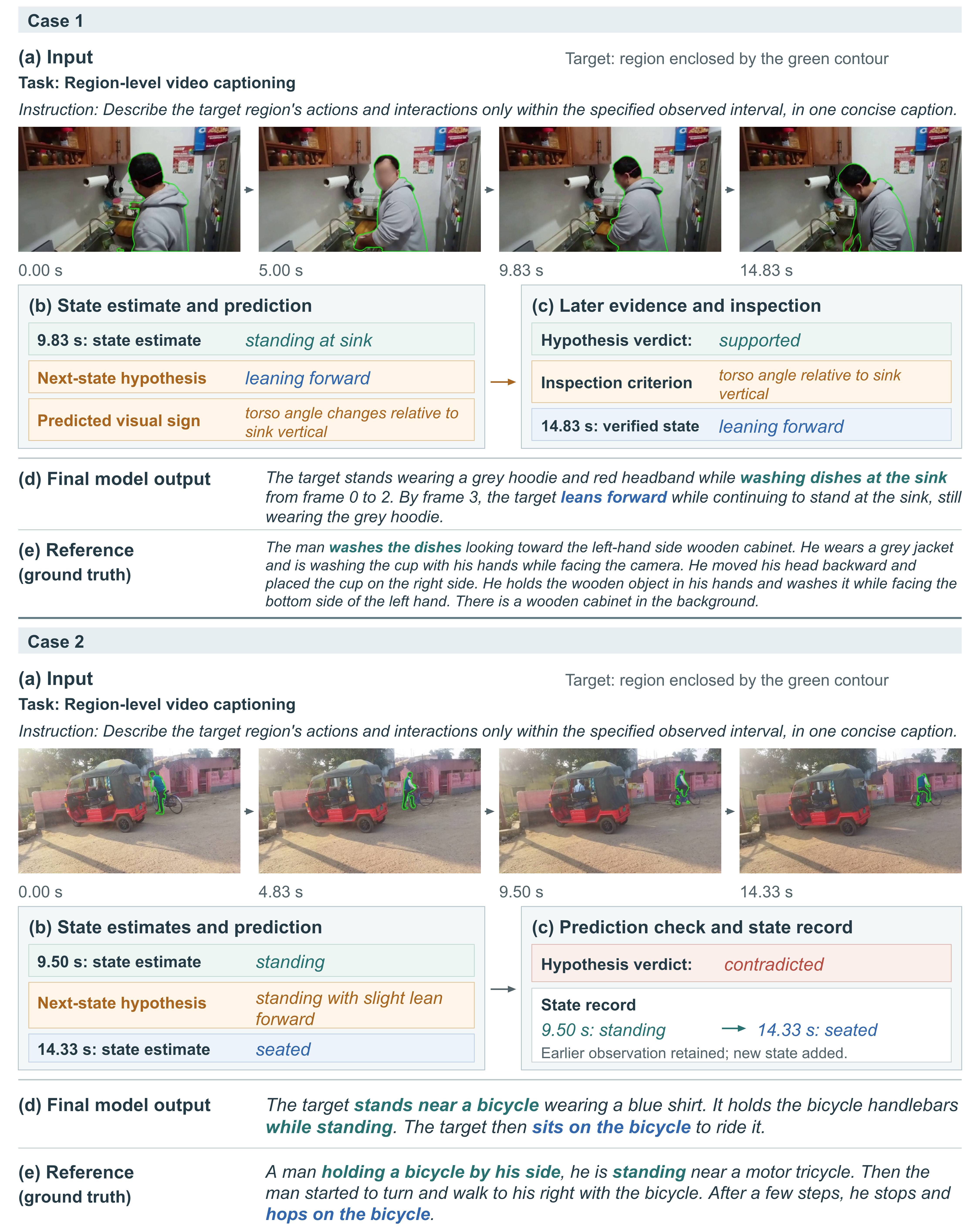}
\caption{Two region-captioning cases illustrating selected posture changes. Each case shows the input frames and target region, an anticipated change, later visual evidence, the model caption, and the reference. Case 1 links an anticipated forward lean to a visual inspection criterion. Case 2 rejects a standing forecast and preserves the observed transition to sitting. Green contours identify the target.}
\label{fig:case}
\vspace{-0.3cm}
\end{figure}

\subsection{Case Study}
\label{sec:case}
We use two region-captioning cases from source videos to illustrate how LASER handles supported and contradicted predictions.
Figure \ref{fig:case} shows the input frames and target regions in panel (a), state estimates and predictions in panel (b), subsequent visual checks in panel (c), and the generated and reference captions in panels (d) and (e).
Green contours identify the target whose state is tracked.

In Case 1, the anticipated forward lean directs inspection toward the target's torso angle relative to the sink.
The later image supports this change, and the final caption describes the lean while preserving the ongoing dishwashing activity.
The reference mentions dishwashing without explicitly describing the lean, so the displayed images provide the evidence for this additional posture detail.
The case illustrates how an anticipated change becomes a specific visual criterion used in the final description.
In Case 2, LASER anticipates continued standing with a slight forward lean, but the later observation shows the target seated on the bicycle.
Sequential visual correction rejects the forecast, retains the earlier standing state, and records the newly observed transition.
The final caption describes the target sitting on the bicycle, consistent with the reference's account of getting onto it.
The case illustrates that an incorrect forecast still leaves room for evidence-supported state readout: subsequent observations determine the transition without changing a supported earlier state.

\section{Related Work}
\label{sec:related}
Object-centered video understanding studies state localization \citep{xue2024learning,soucek2022look}, transformation graphs \citep{sun2025tracking}, and segmentation through appearance changes \citep{yu2023video}.
Region-captioning approaches emphasize object representations \citep{yuan2025videorefer} and local--global context \citep{lian2025describe}, while OSCaR \citep{nguyen2024oscar} and PerceptionLM \citep{cho2025perceptionlm} provide state-change and region-captioning tasks, respectively.
Agent approaches organize reasoning and interaction \citep{yao2023react}, execute visual programs \citep{gupta2023visual}, and gather evidence through iterative inspection \citep{wang2024videoagent}, keyframe selection \citep{wang2025videotree}, or evidence-sufficiency assessment \citep{wang2026active}.
StreamAgent anticipates where and when task-relevant information will appear to guide evidence acquisition and response timing in streaming video question answering \citep{yang2026streamagent}.
Other uses of possible outcomes include counterfactual supervision for video representation learning \citep{kung2025what} and robot behavior forecasting in TRACE \citep{puthumanaillam2025trace}.
Feedback-based approaches revise responses \citep{madaan2023selfrefine} or retain experience as reflections \citep{shinn2023reflexion}, workflows \citep{wang2024agent}, manuals \citep{chen2024automanual}, and evolving contextual playbooks \citep{zhang2026agentic}.
Prompt optimization uses scored candidates \citep{yang2024large}, textual critiques \citep{pryzant2023automatic}, modular instruction search \citep{khattab2023dspy,opsahlong2024optimizing}, and feedback propagation \citep{yuksekgonul2025optimizing}; GEPA uses execution traces to reflect on and evolve prompts \citep{agrawal2026gepa}.
LASER focuses on the role of anticipation in determining what constitutes an informative description of the present: alternative next-step changes identify the object attributes and visual signs to inspect in the current image, whose evidence establishes their values.
Subsequent observations check these values and transitions, while caption discrepancies guide revisions to the responsible attribute-selection, visual-inspection, or expression instructions, making the observation criteria themselves the target of adaptation.

\section{Conclusion}
\label{sec:conclusion}

In this paper, we propose look-ahead state readout (LASER), an anticipation-guided agent that uses possible next-step changes to determine which object attributes deserve attention in the present.
LASER converts anticipated changes into current-state inspection questions and uses subsequent visual evidence to refine the state record for description.
Feedback-guided skill revision improves reusable observation instructions through caption feedback while keeping all model parameters fixed.
Experiments on OSCaR and PLM RCap demonstrate consistent improvements over direct captioning and alternative observation strategies across four MLLM backbones, and LASER outperforms the five compared existing approaches.
Ablations support the importance of inspecting current attributes, identifying distinguishing visual signs, and checking predictions against independent observations.

\bibliography{iclr2027_conference}
\bibliographystyle{iclr2027_conference}

\end{document}